# Towards automating the generation of derivative nouns in Sanskrit by simulating Pāṇini

**Amrith Krishna**
Department of CSE,
Indian Institute of Technology,
Kharagpur, India
amrith@iitkgp.ac.in

**Pawan Goyal**
Department of CSE,
Indian Institute of Technology,
Kharagpur, India
pawang@cse.iitkgp.ernet.in

**Abstract**

About 1115 rules in Aṣṭādhyāyī from *A.4.1.76* to *A.5.4.160* deal with generation of derivative nouns, making it one of the largest topical sections in Aṣṭādhyāyī, called as the Taddhita section owing to the head rule *A.4.1.76*. This section is a systematic arrangement of rules that enumerates various affixes that are used in the derivation under specific semantic relations. We propose a system that automates the process of generation of derivative nouns as per the rules in Aṣṭādhyāyī. The proposed system follows a completely object oriented approach, that models each rule as a class of its own and then groups them as rule groups. The rule groups are decided on the basis of selective grouping of rules by virtue of *anuvṛtti*. The grouping of rules results in an inheritance network of rules which is a directed acyclic graph. Every rule group has a head rule and the head rule notifies all the direct member rules of the group about the environment which contains all the details about data entities, participating in the derivation process. The system implements this mechanism using multilevel inheritance and observer design patterns. The system focuses not only on generation of the desired final form, but also on the correctness of sequence of rules applied to make sure that the derivation has taken place in strict adherence to Aṣṭādhyāyī. The proposed system's design allows to incorporate various conflict resolution methods mentioned in authentic texts and hence the effectiveness of those rules can be validated with the results from the system. We also present cases where we have checked the applicability of the system with the rules which are not specifically applicable to derivation of derivative nouns, in order to see the effectiveness of the proposed schema as a generic system for modeling Aṣṭādhyāyī.

## 1 Introduction

Aṣṭādhyāyī, the central part of Pāṇini's grammar, is a classic and seminal work on descriptive linguistics. Aṣṭādhyāyī provided a complete description of the Sanskrit language spoken at that time period, and is also often praised for the computational principles and programming concepts used in it. Approximately one fourth (about 1115) of rules in Aṣṭādhyāyī deal with generation of derivative nouns (and adjectives), which are derived by affixation from other nouns (or adjectives). The affixes that are used for derivation of derivative nouns are enumerated in the Taddhita section of Aṣṭādhyāyī and hence the affixes that come under this section are called Taddhita. Taddhita section starts with rule *A.4.1.76* तद्धिताः of Aṣṭādhyāyī which is an adhikāra rule and its influence is till the end of fifth chapter i.e. *A.5.4.160* निष्ठवाणिः च. Though Pāṇini did not provide any semantic definition for Taddhita, it is based on rule *A.5.1.5* तस्मै हितम् which means "beneficial to that" (Bhate, 1989), where "that"(तस्मै) is intended for the base nominal from which the derivation will take place. Aṣṭādhyāyī considers both nouns and adjectives as a single category called as Prātipadika (प्रातिपदिक), and hence the affixes used in Taddhita are not category changing affixes (Deo, 2007).

Though Aṣṭādhyāyī was intended for human understanding and usage, our work attempts to automate the process of deriving Taddhitas in complete adherence to Aṣṭādhyāyī. Our work, not only aims to generate the correct final form from any given nominal but also to follow the sequence of rules applied as per Aṣṭādhyāyī in the derivation process. Hence the work requires to automate rules in Aṣṭādhyāyī that deal with Taddhita section as well as other associated rules, which help in the derivation process.

The proposed system adopts a completely object oriented approach in modeling Aṣṭādhyāyī. The rules of Aṣṭādhyāyī are modeled as classes and so is the environment that contains the entities for derivation. The rules are then grouped based on the notion of topicality by virtue of anuvṛtti. In our proposed system the rule group formation is achieved through formation of inheritance network i.e. multilevel inheritance formed between individual rule classes, which has been inspired from the inheritance network that Pāṇini used in Aṣṭādhyāyī (Deo, 2007). Pāṇini uses anuvṛtti to carry forward the inherited components to child rules, though it needs to be noted that signifying of the inheritance is one of the aspects of anuvṛtti, and hence our proposed model does not form the inheritance network over all the usages of anuvṛtti, but rather a subset of it. The principles discussed in Aṣṭādhyāyī that enable rule selection and rule application like *A.1.4.2* विप्रतिषेधे परं कार्यम्, *siddha* and *asiddha* are built right into the core architecture of proposed system. In addition to these principles, the system provides a functionality to adopt different conflict resolution methods that are discussed outside of Aṣṭādhyāyī. The system does not restrict itself by adopting any particular conflict resolution method, instead it facilitates to try out various methods that have been mentioned in various scholarly works. This will thus help to evaluate various methods and report on their accuracy as no single set of conflict resolution methods has gained consensus among the scholars.

The rest of the paper is organized as follows. In Section 2, we will be discussing about various attempts towards formalizing rules of Aṣṭādhyāyi, and modeling Aṣṭādhyāyi in part or full to a automated system. In Section 3, we will look into the linguistic and structural features of Taddhita section. We will be discussing about various linguistic characteristics that the affixes in the domain possess. We will also be looking into the arrangement of rules in Taddhita section and how the arrangement forms an inheritance hierarchy. In Section 4 we will be describing the working of the proposed system, tools and techniques used for the implementation and also modeling of data environment and rule classes. Section 5 will show a step by step derivation of a nominal from another base noun. Section 6 will show the applicability of the proposed schema as a general framework to model entire Aṣṭādhyāyi by considering how the proposed system handles rules that are not specific to the derivation of a derivative noun. Section 7 details about the evaluation framework and results. The section also presents some analysis of incorrect cases and other special instances that we encountered in our system. Finally, Section 8 discusses the bottlenecks in automating the Aṣṭādhyāyi that we encountered during the implementation of our system, salient features and application domains of the system along with directions for future work.

## 2 Related Work

Aṣṭādhyāyi has received much attention from computational linguists from the latter half of 20th century. Aṣṭādhyāyi was much lauded for its brevity, completeness and computational insights it provides. There have been works from eminent scholars about the formalism of Pāṇini's Grammar, its expressive power, and the derivation process (prakriyā) it follows. Seminal works from Cardona (1965) and Staal (1965) on formalizing rules of the grammar with stress on the meta-rules that state about the context sensitive aspects was a starting point with further enhancements from Cardona (1969) where he applies his formalization for more rules that is related to phonetic changes.

Cardona (2009) highlights the relevance of affixation and how it is well integrated with syntax as a continuum in Pāṇini's derivational system. He points out the contrast that Pāṇini's system bears with the system that western grammarians follow, where morphology and syntax are treated as independent components in derivation. Penn and Kiparsky (2012) focus on the expressive power of Aṣṭādhyāyi rules and also the expressive power of formalism that Pāṇini used to design Aṣṭādhyāyi. They demonstrate that the formalism of the grammar has far more expressive power than that of regular languages and Context Free Languages. Their work emphasizes on the power of formalism that has built-in capacity for disambiguation at syntactic level. Hyman (2009) developed a finite state transducer after re-framing individual rules in Aṣṭādhyāyi, resulting in generation of strings that belong to regular languages and performs sandhi (सन्धि) at word boundaries for any two given word combinations. Hyman had introduced an XML vocabulary for encoding rules in Aṣṭādhyāyi that helped him in implementing the finite state transducer. Scharf (2009) discusses about Scharf's and Hyman's combined efforts in developing XML formalization that not only deals with sandhi but also with nominal declensions and verb conjugations.

There have been various attempts to automate Aṣṭādhyāyi in parts as well as modeling it entirely. Goyal et al. (2009) implemented an inflectional morphology generator that takes as input a noun from the user and then generates all 21 forms of noun declensions, known as *vibhakti* system in Sanskrit grammar. The authors talk about the programming perspectives that need to be considered when encoding rules in Aṣṭādhyāyi, and various computational aspects that Aṣṭādhyāyi possesses. They also talk about the need of conflict resolution methods for competing rules that can be applied in the same context. Jha et al. (2009) have developed a system that is an inflectional morphology analyser. They have developed independent systems for verb and noun forms and their corresponding inflections. Though their work was not related to simulating Aṣṭādhyāyi, but they claim that they take into account the Pāṇinian way of analysis. Satuluri and Kulkarni (2014) takes on generation of Sanskrit compounds called as samāsa that deals with about 400 rules of Aṣṭādhyāyi which helps them to form compound words from independent words in Sanskrit. Their work talks about various kinds of semantic features that act as the constraints governing compound formation.

Subbanna and Varakhedi (2009) have talked in length about the Computational Structure of the Aṣṭādhyāyi, and introduced the concept of rules that continuously observe the environment or the subject to which modifications are to be made. They have also talked about *Siddha,Assidhavat and Asiddha* principles used in Aṣṭādhyāyi.There has been an in-depth study on *siddha and asiddha* principles by johshi and Roodbergen (1987), where they talk about the order in which rules need to be applied. Subbanna and Varakhedi (2009) mentioned about the grouping of rules based on the general-exception relation between rules and formation of rules as a tree structure, but commented that the feasibility of automation needs to be checked. They also talk about various conflict resolution methods that are mentioned in the sūtras as well as in other vārttikas. Subbanna and Varakhedi (2010) presented a computational model based on the principle of *asiddhatva*, an improvised model over the one discussed in Subbanna and Varakhedi (2009). Mishra (2009) talks of the nature of grammar which performs the analysis of constituent elements and then its reconstitution using various set of operational rules as mentioned in Aṣṭādhyāyi. Mishra (2010) in his work discusses about *vedāṅga* principles and extends his work by considering the common methodological approach of ancillary disciplines for rule application. His work provides a good walk-through for the entire derivation process that begins with introduction of atomic elements to coming up with the desired final form. Kulkarni (2009) establishes the issue with phonological over-generation that can occur, if one is to strictly adhere to the rules defined by Pāṇini. Jha and Mishra (2009) sheds some light in formalizing semantic categorization rules when he deals with kāraka systems. These kind of issues have been a matter of debate among linguists for quite long. Many principles that are not stated in Aṣṭādhyāyī but in other texts written at various points of time have surfaced to deal with such issues, mainly those which concern about conflict resolution in rule selection. Cardona (1997) discusses about various principles like *utsarga-apāvada*, *antaraṅga-bahiraṅga*, *nitya-anitya* etc. in detail.

When it comes to automating Taddhitas, the Sanskrit Heritage System is an existing system that can recognize taddhitas and perform the analysis, but it does not generate the Taddhitas and only the lexicalized Taddhitas are recognized during the analysis (Goyal and Huet, 2013). To the best of the authors' knowledge, the system which we are going to propose is the first of its kind, that is focused specifically on generation of derivative nouns (Taddhitas). The proposed system embraces a unique approach of forming rule groups where similar rules are grouped together to form a Directed Acyclic Graph (DAG). The similarity of rules is based on the notion of topicality present among the rules by virtue of usage of anuvṛtti. The approach can be treated as analogous to the model, what is proposed in Subbanna and Varakhedi (2009), but they propose the formation of similar topic DAG through utsarga-apāvada relations. One cannot comment on the similarity of the approaches without a proper comparison of DAGs formed from both of them. Moreover Subbanna and Varakhedi (2009) mentioned that they had not checked the feasibility of automating their notion of rule group formation. Another important feature that our system uses is that it automatically notifies the relevant rule classes whenever the data environment state changes. This eliminates the overhead of linear searching over each rule, or each rule polling the environment to find its application. Instead the mechanism allows the rules to be updated about environment state changes whenever there is one and yet the rule classes can refrain from state dependency issues with the environment (Szallies, 1997).

## 3 Linguistic and Structural aspects of Taddhita Section

In the Taddhita section, Pāṇini identifies about 300 semantic relations under which Prātipadikas can be generated with Taddhita affixes. It is mentioned as a sub-section to *pratyayādhikāra* that deals with all kinds of affixes. The rules in Taddhita section deal with three entities namely semantic relations, affixes and stems or collection of stems from *gaṇapāṭha*. The rules are defined in such a way so as to facilitate affixation of the proper Taddhita affix with respect to semantic relation intended. The rules often deal with properties of the entities involved at various levels from phonological, morphological, syntactic and semantic levels. There are two types of derivations possible that involve Taddhita affixes (Sharma, 2002).

<div align="center">Prātipadika + Taddhita-affix</div>
<div align="center">Prātipadika + Taddhita-affix + strī-affix</div>

### 3.1 Linguistic Phenomena in Derivational Morphology

Derivational morphology in Sanskrit, like in many other languages poses some of the well known facts about many to many correspondences between forms and affixes. Taddhita affixes exhibit affix polysemy, homonymy, synonymy and non-compositionality. In affix polysemy, the same affix is used to denote related senses like in the case of patronymic and provenance relation. In affix homonymy the same affix is used in distinct and unrelated semantic contexts like in the case of personal nouns or abstract nouns. Affix synonymy deals with the same semantic sense but uses different affixes. For example, for the patronymic relation अपत्यम्, multiple affixes can be used, like अ(ण), अयन(फक्), इ(ञ्) depending on the stems used (Deo, 2007).

### 3.2 Organization of Taddhita rules

Taddhita section is primarily subdivided into five *pratyayādhikāras* or domain of control of five pratyayas.

- ■ The five rules are:

*I*  *A.4.1.83* प्राक् दीव्यतः अण् - अण् suffix is the default affix to be used for affixation for all the rules till *A.4.4.1*. The influence of *A.4.1.83* is till the term दीव्यति is found (or till another pratyayādhikāra is found) and दीव्यति appears in rule *A.4.4.2* तेन दीव्यति खनति जयति जितम्.

*II*  *A.4.4.1* प्राक् वहतेः ठक् । - ठक् suffix is the default affix to be used for affixation for all the rules till *A.4.4.76*. वहति appears in rule *A.4.4.76* तत् वहति रथयुगप्रासङ्गम्.

*III*  *4.4.75* प्राक् हितात् यत् - यत् is the default affix to be used for affixation for all the rules till *A.5.1.5*. हितम् appears in rule *5.1.5* तस्मै हितम्.

*IV*  *5.1.1* प्राक् क्रीतात् छः - छः is the default affix to be used for affixation for all the rules till *A.5.1.37* क्रीतम् appears in rule *5.1.37* तेन क्रीतम्.

*V*  *5.1.18* प्राग् वतेः ठञ् - ठञ् is the default affix to be used for affixation for all the rules till *A.5.1.115*. वतिः appears in rule *5.1.115* तेन तुल्यं क्रिया चेत् वतिः .

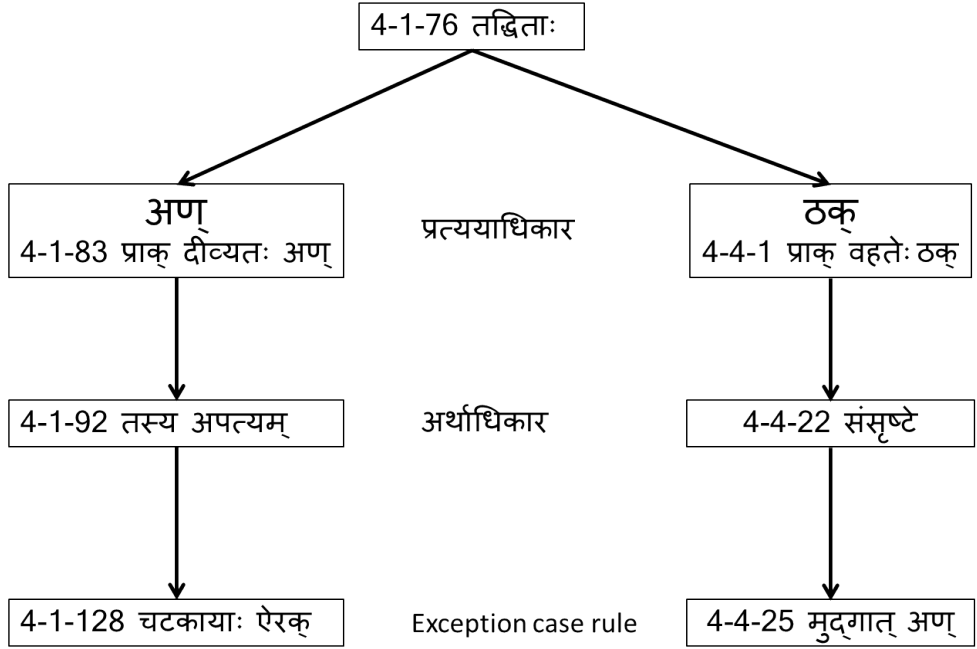

Figure 1: An instance of inheritance hierarchy in Taddhita section

Now each of the above given rules, which can be called as the pratyayādhikāra rules, specifies what is the most prominent affix or the default case affix that can be applied to all the rules which are under its domain. For the set of rules under a single pratyayādhikāra, they are further categorized on the basis of arthādhikāra rules. Each arthādhikāra heads a set of rules which are a proper subset of rules that come under the pratyayādhikāra. The arthādhikāra rules state the semantic conditions or the semantic rules under which the default affix can be attached. Arthādhikāra rules serve the purpose of a topic head that groups a set of rules as a rule group based on a topic, which is a semantic relation here and it also acts as an operational rule (विधि), as it carries the semantic sense under which the affix can be attached. Rules *A.4.1.92* तस्य अपत्यम्, *A.4.2.1* तेन रक्तं रागात्, *A.4.2.69* तस्य निवासः etc. denote the semantic senses such as patronymic, "coloured by means of that" and provenance respectively. This kind of arrangement handles affix polysemy and affix homonymy. But to handle affix synonymy, there are other operational rules, that come under the domain of arthādhikāra rules. They can be seen as exception rules or rules to handle special cases. These rules limit the application of default affix specified in pratyayādhikāra and mention what other affixes can be used instead, under special conditions. In this way, affix synonymy and non-compositionality is taken care of.

Deo (2007) calls this form of arrangement as constrained separationism, where the rules form a multilevel single inheritance network. The template for the multilevel inheritance network will be of the type, "Default Affix rule" -"Semantic Sense rule" - "Special case rules". Figure 1 shows an instance of how the hierarchy in Taddhita works. As already discussed, *A.4.1.83* states about the default affix अण् for rules under its domain, *A.4.1.92* states about the patronymic relation under which a nominal will get the default affix. Now rule *A.4.1.128* states about a special case when the stem चटका is used with the patronymic relation. In such a case the suffix ऐरक् needs to be attached. This rule overrides the default case rule. Now consider rule *A.4.4.25*. This rule talks about usage of अण् as an exception or special case when a stem मुद्र is used. The rule has *A.4.4.1* as its default affix rule and the default affix for the domain is ठक्. The rule's semantic sense is संसृष्टे, which means 'properly mixed with'. Here the rule *A.4.4.25* accounts for the non-compositionality of the affix अण् and is not specified in its domain but as an exception rule in another default affix domain.

It is to be noted that there are some more rules in the Taddhita section which do not come under any of the five pratyayadhikāra rules. Those rules have arthādhikāras of their own but has no default affix to be attached. This group of rules is treated as extraneous to otherwise systematic network of the section (Bhate, 1989). As we will be discussing about the implementation schema of proposed generation system in the subsequent section, it will become evident that this anomaly in no way is going to affect the system.

## 4 Implementation

### 4.1 Derivational Process

In order to discuss about the derivation process or (प्रक्रिया) of a derivative noun, let us take the derivation of derivative noun औपगव from the nominal उपगु, which can be considered as the "Hello world" in Taddhita section. The essential steps in the derivation process are shown in Table 1. औपगव which means son of upagu, comes from the semantic sense अपत्यं by the rule 4.1.92 तस्य अपत्यम्, and the correct Taddhita affix अण् is introduced to the environment. Once the Taddhita affix is introduced, what remains is a series of operations on the environment that results in a final form. The rules so triggered may be seen as a continuous iteration of two sub-processes. One is identification and assignment of technical terms (संज्ञा) to the environment string or to a relevant substring of it. By this, the environment does not get modified but gains the technical term as an attribute. The interpretive rules assign the technical terms to the environment and those rules are shown in the first column of Table 1. Now, by virtue of these attributes, the environment gets modified through operational rules that come within the domain of attributes as shown in the third column of Table 1. The effect of the rule on the environment can be seen in the second column. Triggering of operational rules is the second sub-process in the iteration and the derivation process stops when no more attributes can be attributed to the environment. Ideally, by that time the desired form must be derived.

### 4.2 Overview of the Implementation System

For automating the derivation process, we suggest the following method which is based on object oriented concepts. Each rule forms a class and each instance of the rule class (henceforth to be referenced as rule itself) is registered with the environment, such that whenever there is a change in the environment, the rules are notified. Each rule checks for the possibility of it being applied on the environment and for a rule, if all its conditions are satisfied, ideally the rule can be applied on the environment. However, in a general scenario multiple rules may claim their competency for application on the environment. To handle such scenarios we keep those competing rules in a list called candidate list. Then a conflict resolution method is employed which decides the winner rule. The winner rule gets to apply on the environment and other rules are removed from list. By removal of rules other than the winner rule, we mean that the removed ones are not applied on the present instance of environment, although they are notified when the environment change happens again. Figure 2 shows the schema of the implementation

| Interpretational terms to be assigned | Derivation environment | Operation rule to be applied |
|---|---|---|
| | उपगोः अपत्यं | |
| | उपगु ङस् अपत्यं | |
| तद्धित (4.1.76 तद्धिताः) | | |
| | उपगु ङस् अण् | 4.1.92 तस्य अपत्यम् |
| इत् (1.3.3 हलन्त्यम्) | | |
| | उपगु ङस् अ | 1.3.9 तस्य लोपः |
| प्रातिपदिकम् (1.2.46 कृत्तद्धितसमासाः च) | | |
| | उपगु अ | 2.4.71 सुपः धातुप्रातिपदिकयोः |
| अङ्गम् (1.4.13 यस्मात् प्रत्ययविधिः तदादि प्रत्यये अङ्गम्) | | |
| | औपगु अ | 7.2.117 तद्धितेषु अचाम् |
| भ (1.4.18 यचि भम्) | | |
| | औपगो अ | 6.4.146 ओः गुणः |
| संहिता (1.4.109 परः सन्निकर्षः संहिता) | | |
| | औपगव् अ | 6.1.78 एचः अयवायावः |
| | औपगव | |

Table 1: Derivation Process for औपगव. Here each horizontal line partition represents one step in the derivation, where one of the operations among insertion, elision or substitution takes place in column 2. Column 1 shows assignment of some technical term, and column 3 shows subsequent operation after gaining the technical term as a property. Column 2 shows the effect due to the operation. In column 2 we can see the effect of the rules on the environment, where an insertion, elision or substitution takes place.

system.

As it is evident from discussion about Taddhita section, Pāṇini among various applications of anuvṛtti, uses it for carrying the topicality between different rules as well. It is to be noted that adhikāra rules are rules whose sole purpose is mentioning the topicality of the rules under its domain of influence. But even for adhikāra rules, anuvṛtti itself is used for carrying the domain's influence to other rules. Apart from the adhikāra rules, there are other rules as well in which anuvṛtti is used to carry the topicality. For example consider the rules concerning इत्, i.e. rules from 1.3.2 to 1.3.8. Here उपदेशे and इत् are being carried forward to other rules as well. उपदेशे, which means 'when an upadeśa is encountered', perform some action. This condition leads to a common topicality for all the rules under the anuvṛtti. This is similar to the notion of arthādhikāra rules in Taddhita where the notion of semantic sense is being carried forward to subsequent rules through anuvṛtti. Both the techniques, anuvṛtti and adhikāra are employed in the entire Aṣṭādhyāyī and are not unique to the Taddhita section. We can infer that a subset of the anuvṛtti rules, mostly the ones which carry the notion of topicality can be used to design an inheritance hierarchy of classes. For the implementation we will be using this notion of topicality via anuvṛtti, in grouping of rules to form rule groups which is an inheritance hi-erarchy among rules and the rule in which the portion of anuvṛtti appears, becomes the head rule.

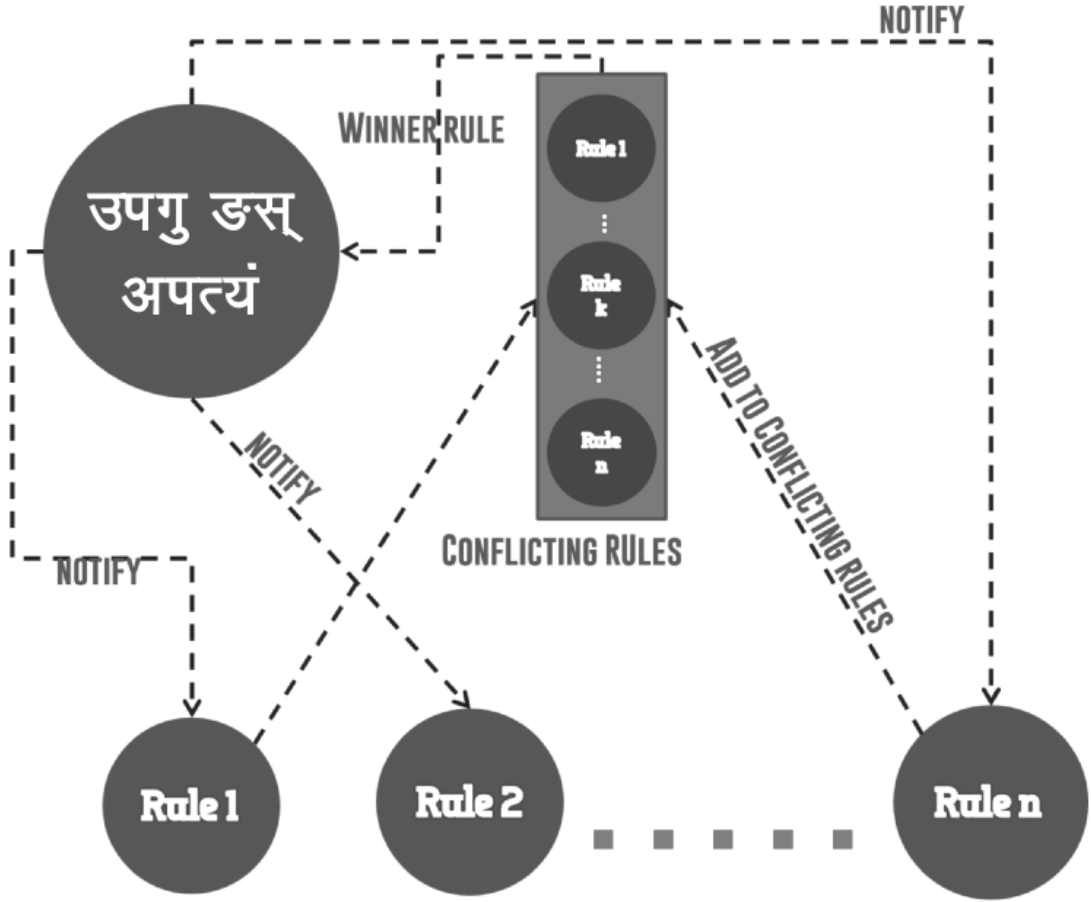

Figure 2: Overview of Implementation System

### 4.3 Tools and Techniques Used

We are following a completely object oriented approach for the implementation of the system. The following tools and techniques will be employed to achieve the principles discussed in section 4.2

#### 4.3.1 Observer Design Patern

Observer design pattern defines a one-to-many dependency between objects so that when one object changes state, all its dependents are notified and updated automatically (Gamma et al., 1994). The object on which the state changes occur is called subject and it maintains a list of its dependents, called observers. As shown in Figure 3, each observer object which is the sūtra or rule in our case, is registered to an object called as "Subject" class which represents the environment in our case. Subject class has a method to register the observers. Whenever a change in value of some attribute of subject occurs, it calls the method 'notify' of observer abstract class, which is implemented for each object of the Observer.

#### 4.3.2 Multilevel Inheritance

To form rule groups i.e. inheritance network among rules we use multilevel inheritance between the rules. Though multilevel inheritance is allowed, multiple inheritance is not allowed in the system and hence no single rule will inherit from two distinct rules directly. Figure 4 shows the inheritance achieved for each class in Aṣṭādhyāyī. All classes inherit from a base class called "sūtra", which is a generic class that defines all possible features that a rule or sūtra can possess. All other rules inherit from it. The adhikāra rules, anuvṛtti of interpretive rules, default condition rules etc. form super class for other operational rule classes.

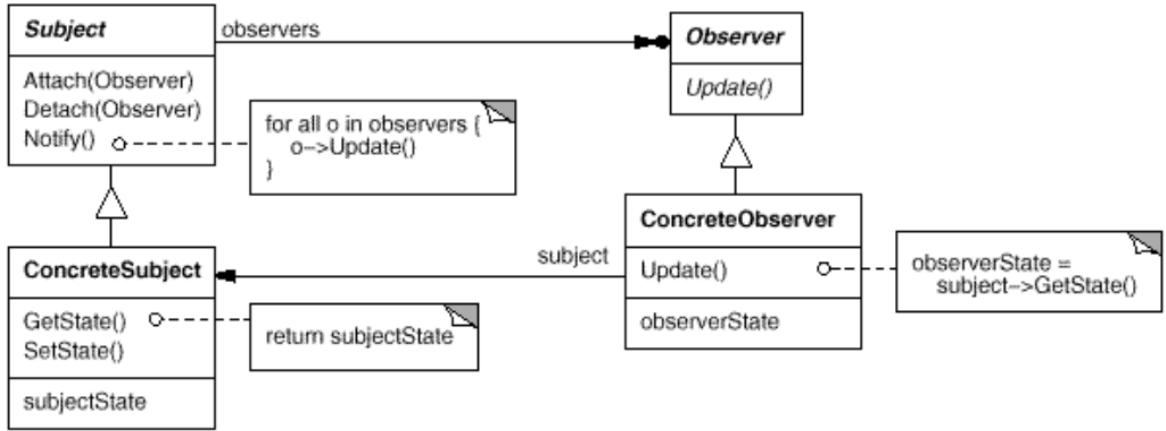

Figure 3: UML diagram for observer design pattern

### 4.4 Rule Triggering and Propagation

The environment forms the subject class for the observer design pattern. However, not all rules are observing the environment but only those rules that do not come under the domain of any adhikāra rules or a controlling head of a rule group through anuvṛtti (which are mostly rules that assigns the technical terms to assignments). Those are represented by classes R1, R2 and so on in Figure 5. Then there are rule groups as represented by RG1, RG2 and so on. These are collections of rules with same head. The rule classes that come under the same head in the group, inherit from the head rule class. The head rule object, as per observer design pattern registers the inherited classes' objects as observers. Now the head class notifies the observers whenever the environment satisfies the head rule's conditions. The conditions checked are those conditions that need to be satisfied by all the rules registered under the head. Here, by 'head', we mean either an adhikāra or a component passed on by anuvṛtti. Now top level rules are those rules which observe the environment directly and get notified whenever the environment changes. For a rule, if the environment satisfies its conditions, it will be added to the candidate list. But if the environment satisfies the conditions that is applicable to an entire rule group, then the environment object is passed on to the next level and this continues till an exception or specific rule is encountered or else returns back to where the default rule resides. By this model, we can employ conflict resolution at each level, and resolve some of them locally and only rules that have no common head at any level come to the candidate list at top-most level, from where the winner rule will be selected.

Once a winner rule is selected, the rule's intended action is executed first and then its parent object is called which performs its relevant portion in execution, if any. This continues until all the rules in hierarchy are called. It is to be noted that many a times certain rules like the adhikāra rules do not have anything to execute of their own; in such cases the rule object just passes on the environment to its parent object. By this design, redundancy of hard coding the same rules again and again per rule object is saved, just like the way anuvṛtti helps a person to avoid repeating the rules when reciting Aṣṭādhyāyī.

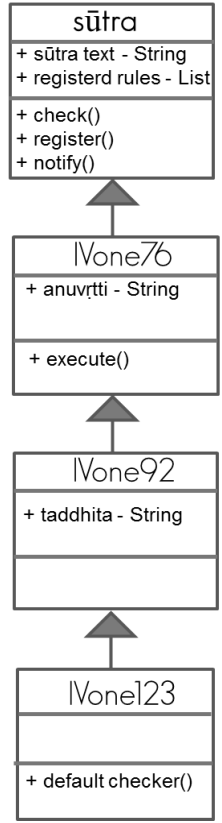

Figure 4: UML diagram for multilevel inheritance

### 4.4.1 Working

Let us consider the working of the system with reference to affixation for semantic relation अपत्यम्, i.e. the patronymic relation. Figure 6 shows the trace for the affixation process of Taddhita affix when the semantic relation is patronymic and the stem is चटका. In Figure 6, the solid lines show inheritance hierarchy between the rule classes. The dashed line on the left of each block shows traversing of the hierarchy by triggering of rules from head to specific rules. The traversal checks for eligible rule to apply on environment within the rule group, and this can be called as the checking phase. The dotted lines on the right of each block show the trace of rules that get executed. The starting node, i.e. the node that heads the dotted edge with label 1 is the exact rule that gets executed and other nodes in the path show those rules that have been executed due to anuvrtti. This can be called as the execution phase. As an example let us consider the case where the environment is initialized with चटकायाः अपत्यम्. The triggering starts from head rule at the top level i.e. from *A.3.1.1* to rule *A.4.1.128* as shown in Figure 3. Here in rule *A.3.1.1*, it does not have any condition to check, so it directly notifies all of its direct descendants. Now among the direct descendants, rule *A.4.1.1* checks for presence of any of the two affixes ङि, आप् or if the environment has a prātipadika in it, i.e. it checks for conditions mentioned directly in the rule. As it is a prātipadika, the condition will evaluate to 'true',and all its descendants are notified. In due course, *A.4.1.76*, *A.4.1.83* are also notified. These rules as well do not have any extra checks as they are adhikāra rules and hence all its direct descendants are notified. When *A.4.1.92* is notified, it checks for semantic condition and the checking turns out to be true for *A.4.1.92*, while the checking will evaluate to 'false' for all of its sister nodes, i.e. other direct descendants of *A.4.1.83*. The special case rules registered under *A.4.1.92* are notified, of which *A.4.1.128* satisfies the remaining conditions. As it does not have any rules registered to it, it becomes the terminal node and hence it is added to the candidate list. Since for this case no other rule is contesting, the rule emerges winner and starts its execution from *A.4.1.128*.

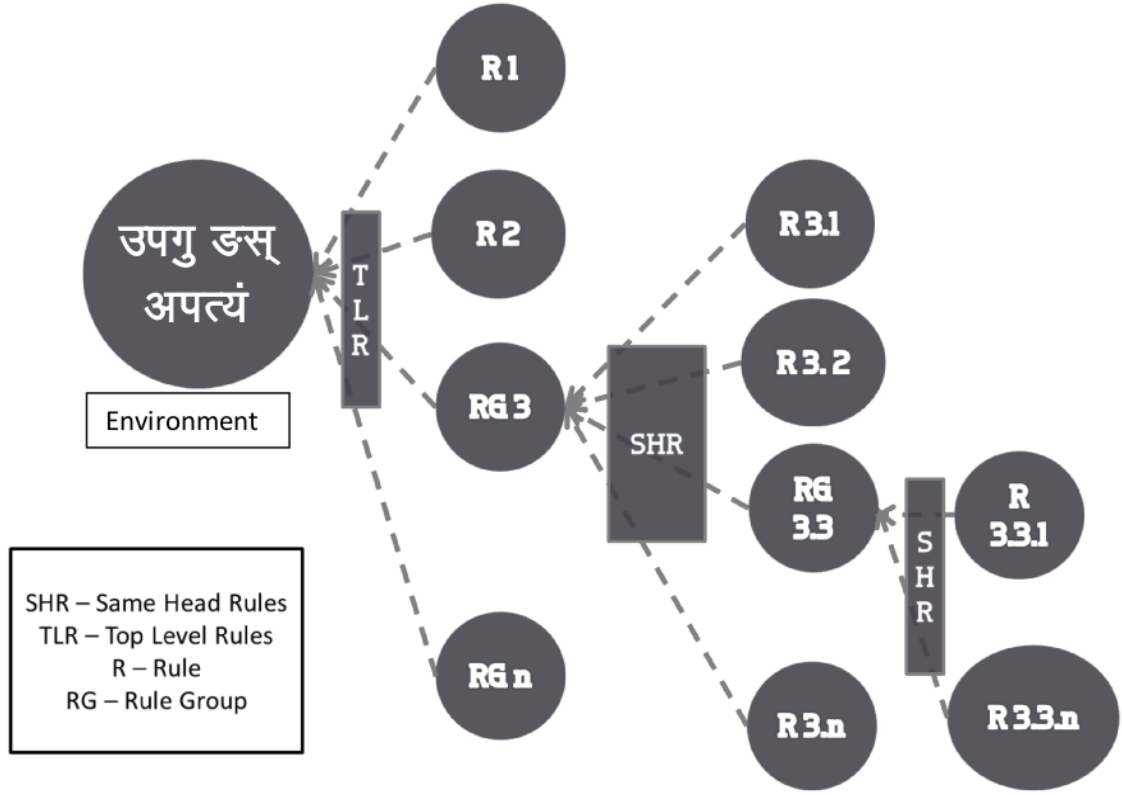

Figure 5: Triggering schema.

The rule adds the affix ऐरक् to environment, and passes the environment to its parent class i.e. *A.4.1.92*. *A.4.1.92* does not have anything to execute of its own, so it simply passes environment to *A.4.1.83*, which checks if any affix is already attached, as the affix 'ऐरक्' is attached in this case, no action is performed. The environment gets passed to parent node of each rule finally this terminates at top level rule *A.3.1.1*. Now consider the derivation of उपगोः अपत्यं. As with the case of चटकायाः अपत्यं, the path of the rules checked for eligibility of rule application remains the same till rule *A.4.1.92* is reached. Once *A.4.1.92* is reached it will notify all of its descendants as well. But since no rule will find its application, *A.4.1.92* will become the final node of the path here. It is added to candidate list, which while executing will simply call the parent rule *A.4.1.83*. *A.4.1.83* checks for presence of any new Taddhita affix in the environment. If the check evaluates to false, i.e. if no new Taddhita affix is found, *A.4.1.83* treats this as default case scenario and attaches अण्, the default case affix to environment. After each execution of a rule, the system checks for presence of any new assignment of 'technical term' or else all rules in the top level are notified as discussed in Section 4.1.

The arthādhikāra rules inherit from its corresponding pratyayādhikāra rules, apart from the ones already mentioned in Section 3.2. When it comes to affixation, during the checking for eligibility of a rule to apply i.e at the checking phase, we need to traverse the pratyayādhikāra rule, before an arthādhikāra rule is reached. Though a pratyayādhikāra rule is visited during checking, no action is taken there. The pratyayādhikāra rule simply passes the environment to all arthādhikāra rules which are its direct descendants. In fact for a single affixation, all the pratyayādhikāra rules get notified from its parent rule, and those rules in turn notify all their direct descendants as there is not enough information to select a single pratyayādhikāra at during the checking phase. So in effect, the process of affixation for taddhita starts by checking for the right semantic condition i.e at the arthādhikāra rules as it is in the case of traditional system of derivation. Before that, the other rules either simply pass on the environment to their descendants or check for conditions that is necessary for the process to qualify as a case for affixation under taddhita. The effect of pratyayādhikāra rule comes during the execution phase i.e at the applying of affix phase and not on the checking phase. During the execution phase the pratyayādhikāra which is parent to the winner arthādhikāra rule acts as the final gate that makes sure that the environment has the valid taddhita affix added to the environment before it reaches its parent. The rule checks if any taddhita affix is introduced by virtue of special case rules, and if no such execution has taken place, then the default affix is added to the environment. The environment is then passed on to higher level rules that takes care of other generic aspects about the environment.

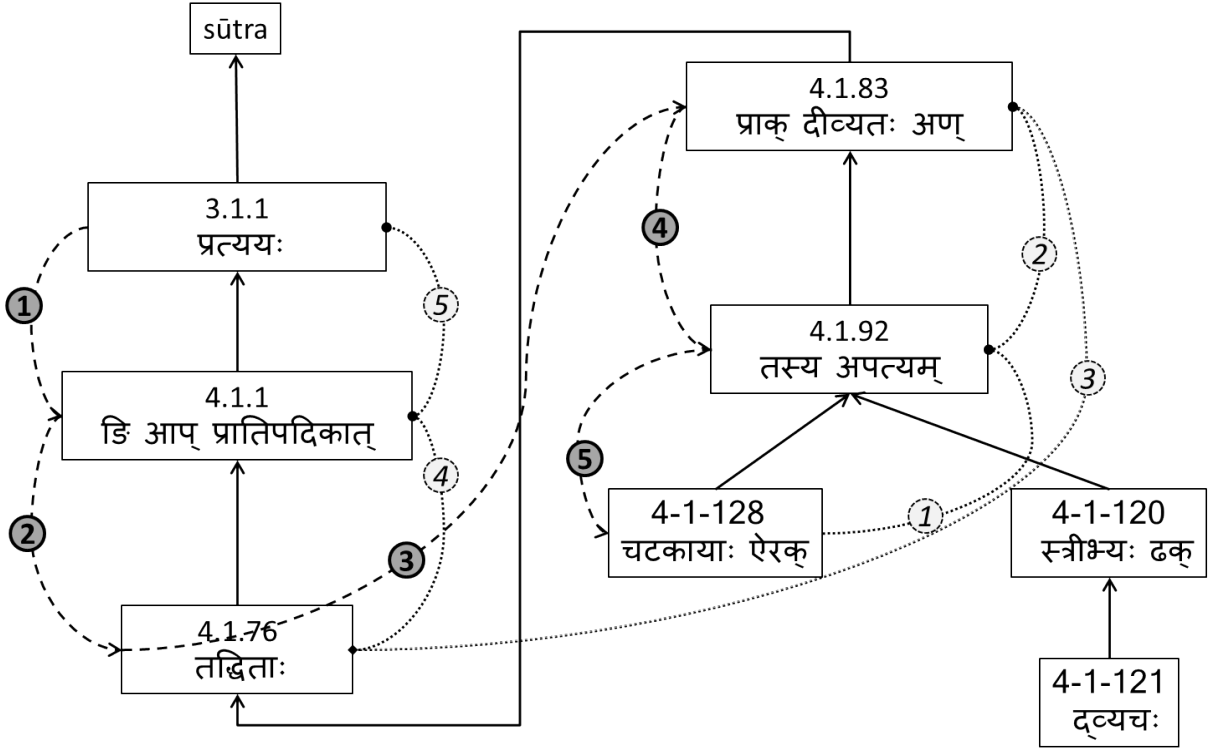

Figure 6: Affixation under patronymic relation for चटका

## 4.5 Environment Representation

The environment is an object which contains the data entities that take part in the derivation process. Entities can be stems, affixes, augments, characters or any of their properties. The most basic and atomic entity in the environment is another object called 'śabda'. The environment also contains various instances of the class 'śabda collection', which contains a sequence of references to śabda objects along with a set of attributes that belongs to the collection. Instances of 'śabda collection' are used to represent various technical terms that may be attributed to environment or a part of it. Figure 7 shows how the environment is set up. Figure 7 shows the environment after the affixation of Taddhita affix अण्. As already seen in Section 4.4.1, the Taddhita affixation happened due to presence of technical term prātipadika. Now the technical term prātipadika is modeled as an attribute of the environment object, and it is an object itself of the class 'śabda collection'. It contains references to sequence of all the śabda objects, which is collectively eligible for the technical term prātipadika. Similarly inside 'pratyaya' object, it has an attribute 'it marker' which also is an object of 'śabda collection' class called 'it'. If you notice though the *it* (इत्) marker is ण् in अण्, it is not the śabda's property that it is an *it* marker. It is the property of pratyaya that gives the śabda ण्, the property of *it marker*. This notion is captured very well in the system. It is essential that we store them as attributes separately for further reference and usage in the derivation process. For example, consider the derivation process for āśvalāyana. The term āśvalāyana is formed from aśvala by affixing 'phak' pratyaya. Here the 'it' marker is 'k', which will be stored in 'it marker' object, and later it will be elided, and hence be removed from the 'text value' of pratyaya object by application of rules *A.1.3.3 and A.1.3.9* . In due course of derivation the pratyaya object will get a substitution of 'āyana' for the remaining 'pha' by rule *A.7.1.2.* Now in order to complete the derivation process, rule *A.7.2.118* should stand valid in one of the subsequent steps. Rule*A.7.2.118* requires a Taddhita affix with k as 'it' marker. If we had not stored this information as a separate attribute earlier, we would have lost this information and derivation would not have completed.

In case of *asiddhavat* rules as discussed in Subbanna and Varakhedi (2009), the environment makes a complete copy of itself; one object is used for checking the conditions while in the other object, all the operations are applied. Once the system returns back to siddha section, the copy used for checking the conditions is discarded. It is also to be noted that in the representation the space between entities are also śabda objects, representing the virāma (**विराम**) as per the rule *A.1.4.110* विरामः अवसानम्.

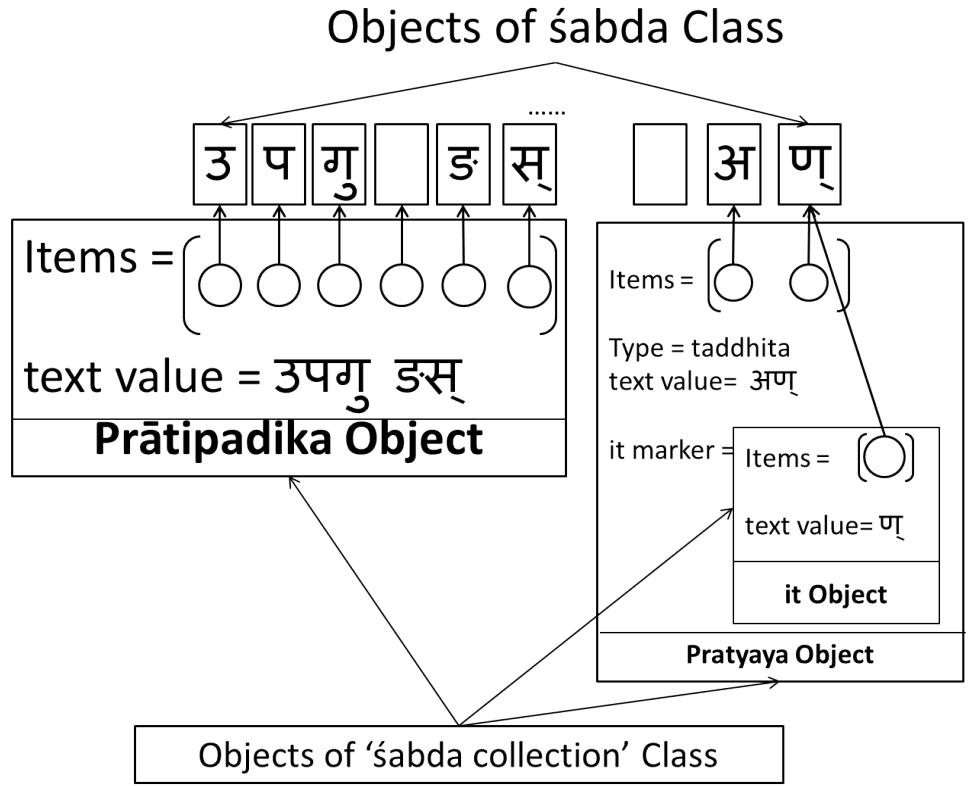

Figure 7: Representation of śabdarūpa object

### 4.6 Rule Selection and Conflict Resolution

There are instances in Aṣṭādhyāyi where multiple rules, which we call as candidate rules, find their suitability to be applied on the derivation environment. In such cases, we need to employ mechanisms that help us in resolving the conflicts. There is little information about conflict resolution among candidate rules in Aṣṭādhyāyi, though commentaries on Aṣṭādhyāyi by other scholars do mention about such mechanisms. It may be the case that Pāṇini had assumed those principles which were prevalent in his time as a prerequisite to understand his treatise on grammar. But this has led to debates among the scholars resulting in different schools of thought. In the wake of such a scenario, we have decided to implement the conflict resolution module as a separate pluggable module in our system, so that we can try out different methods on which the scholars in general have come to a consensus. But our system internalizes one concept as a default standard, the rule *A.1.4.2* विप्रतिषेधे परं कार्यम्. The rule is discussed in Aṣṭādhyāyi itself and the details of how it works is discussed in Section 6. From various commentaries on Sanskrit grammar, we have also considered the mechanism described in 'paribhāṣenduśekhara' of 'Nāgeśa' as 'pūrvaparanityāntaraṅgāpavādānāmuttarottaram balīyaḥ' which is generally accepted among the scholsrs. This gives the following linear order.

> prior (pūrva) < subsequent (para) < obligatory (nitya) < internally conditioned (antaraṅga) < exception (apavāda)

As it is evident from our discussions, we have centered our system design based on the notion of topicality, and the multilevel inheritance network is formed on the basis of topical head rules and the corresponding child nodes. This makes it possible for us to adopt the utsarga-apavāda principle of conflict resolution without any extra efforts. But when it comes to rules which do not fall under similar topical heads, it can be observed that sometimes proper rule selection does not take place. There are some instances in Taddhita section itself, where utsarga-apavāda principle is not sufficient to resolve rule selection conflict . To resolve such a scenario, we have applied the specificity hierarchy as mentioned in Scharf (2010). Specificity hierarchy deals with a priority wise ordering of rules based on the linguistic features present in the rule (or a combination of the same, if multiple entities are present) from the most concrete to the most abstract features. The linear ordering is as follows:

> Phonetics < Phonology < Morphological < Semantic

The specificity of a rule is determined by the individual specificity aspects of entities that are present in a rule as pre-conditions to be satisfied. It is also to be noted that within the entities with the same specificity, finer granularity may be present. For instance, in the apatya semantic sense in the Taddhita section, there are rules that state conditions for the presence of semantic sense of three varying degree of specificity, *apatyam, gotra and yuvām*, and it goes without saying that all the three fall under the same specificity hierarchy of 'semantic'. But amongst these, gotra is more specific than apatyam, and yuvām is more specific than gotra. Hence when two rules find their application in an environment, where one has apatyam as semantic sense and the other has gotra as the semantic sense, the one with gotra specification will emerge as the winner rule even if the former rule comes after the later rule as per the Aṣṭādhyāyi sūtra ordering. Just like in the case of topicality, there are no explicit markings available in Aṣṭādhyāyi to identify the specificity. We need to encode the same in our rule classes explicitly, similar to the treatment given to topicality in our implementation. We have implemented specificity hierarchy for the apatya section in our prototype system and that has improved the results substantially which will be discussed in Section 7.

We will be discussing the effect of applying these conflict resolution methods in Taddhita section and in other specific instances which are outside the scope of Taddhita in Section 6, to establish the system's effectiveness as an automated system for simulation of the entire Aṣṭādhyāyi.

## 5 Derivation of a Derivative Noun

In this section we will show how the nominal stem औपगव is derived in the system. In section 4.4.1, the affixation is already shown. However, some of the finer details regarding post processing after execution of the rule are not discussed, which we will be doing in this step by step walk-through of the derivation. Please refer to Table 1 for state of environment after each rule is applied.

- ■ Affixation as shown in section 4.4.1. Here the user input "उपगु ङस् अपत्यं", has led to affixation of desired affix उपगु ङस् अण्. As the affix got added to the environment in the form of

'pratyaya' object as shown in Figure 7, two more attributes were also added to the 'pratyaya' object. The two attributes are 'upadeśa' and 'Taddhita'. Contrary to as discussed in section 4.4.1, the attributes are not objects by themselves. This is because the pratyaya object bears reference to the exact sequence of śabda objects as these two attributes and hence separate object instantiation is not required. The prātipadika object which is an attribute of the environment, but signifying the base nominal उपगु ङस्, i.e. the stem upagu in genitive case, goes to "processed" state.

- ∎ Since there are two new attributes that are not yet in processed state in one of the objects of environment, instead of notifying the top level rules, system takes in the attribute upadeśa which is a technical term and is assigned to pratyaya object, and triggers corresponding rule group (in this case the rule group headed by *A 1.3.2*). This leads to instantiation of 'it marker' object as shown in Figure 7 and subsequently to rule *1.3.9* that leads to elision of the *it* marker. Though the marker is elided, neither the object reference nor the śabda object, the marker is referring to, is removed from the environment. In fact the śabda object keeps the marker information (णित्) as a separate attribute.

- ∎ Now since no more rules can be triggered automatically, the system notifies all the top level rules of which, top level rule *A.1.2.46* finds its eligibility due to presence of attribute 'Taddhita'. Though *A.1.2.46* gets Prātipadikam प्रातिपदिकम् from *A.1.2.45* as anuvṛtti, still it is not modeled as a descendant of *A.1.2.45*, as it does not represent topicality or common condition. The effect of the anuvṛtti passed on here is that of assigning of the term, which is an effect on the rule, not a cause or condition on the rule. Environment gets a new attribute Prātipadika.

- ∎ No specific rule group can be invoked by Prātipadika attribute. All top level rules are notified, of which only *A.2.4.71* finds its eligibility, which also happens to be a top level rule. This results in removal of ङस्

- ∎ Similarly, the system will get the object '*aṅga*' (for the technical term aṅga) as an attribute to environment, after the rule *A.1.4.13* find its eligibility, and the system will directly notify the rule group headed by *A.6.4.1* अङ्गस्य. *A.7.2.117* will find its eligibility to apply. The exact same steps are executed for subsequent operations, where the technical terms भ and संहिता are assigned due to the rules *A.1.4.18* and *A.1.4.109* respectively and by virtue of those attributes to environment, rules *A.6.4.146* and *A.6.1.78* respectively are executed resulting in the desired final form औपगव.

# 6  The Schema as a General Schema for Modelling Aṣṭādhyāyī

In section 5, we have seen the working of the system. In this section we will be considering rules that are outside of Taddhita section. Scharf (2010) discusses five cases in which conflict of application occurs between two competing rules. In his paper, Scharf talks about conflict between rules *A.6.1.87* and *A.6.1.88* where the domain of application of *A.6.1.88* is properly contained within the domain of application of rule *A.6.1.87*. Here *A.6.1.88* should emerge as the winner rule, or else 6.1.88 will never be applied to any context. In our system *A.6.1.87* is a direct descendant of *A.6.1.84* which is a direct descendant of *A.6.1.77*. The only aspect that is passed from *A.6.1.77* as part of inheritance is अचि, which is nothing but checking for a vowel as a right context. One might argue that *A.6.1.84* does not have any relevance for this checking, but *A.6.1.84* is an adhikāra rule and hence the execution will never stop at the rule; instead it will surely traverse down to one of the descendant rules. Also, its domain of influence is completely within the set of rules which is under the influence of rule *A.6.1.77*. Now *A.6.1.87* is one such rule. It is also evident from this instance as to how the rule is inheriting a set of mutually exclusive features and conditions by multilevel inheritance, avoiding the need for multiple inheritance. *A.6.1.88* is also one such rule, but it needs an additional checking of condition which is checked also at *A.6.1.87*. So in our system *A.6.1.88* inherits from *A.6.1.87* and becomes descendant of *6.1.87* by virtue of the anuvṛtti of आद्. So whenever environment object is passed on to *A.6.1.87*, it first checks presence of अ or आ as the left context and then notifies all its registered rules including *A.6.1.88*. If the rule is applicable to *A.6.1.88*, i.e., if the environment satisfies all the conditions as demanded by *6.1.88* then it gets activated. So in addition to check for left context as per *A.6.1.87*, *A.6.1.88* also checks for एच् as right context, which is applicable only to itself. This check is further restricting the scope of application to what occurred at rule *A.6.1.77* where the checking was for अच्. If this evaluates to 'true', it will block the direct application of *A.6.1.87* i.e. guṇa will not happen there, but instead vṛddhi will take place. If no rules turn out to be eligible for application, then only *A.6.1.87* will perform the application of guṇa over the environment. Please note that the checking of other conditions that are applicable to these rules are not discussed here as those conditions are obtained by virtue of anuvṛtti and are being checked in the parent rules.

Now let us consider a case where partial blocking occurs when the rules *A.6.1.77* and *A.6.1.101* are in conflict. Our system deals with this in the same manner as we dealt with simple blocking. Rule *A.6.1.101* inherits अचि, i.e. element of *ac* pratyāhāra, as right context. One thing to notice here is that the rule *A.6.1.77* also checks for इकः, i.e., element of *ik* pratyāhāra in the context. But this is not carried forward as anuvṛtti. Now this rule is modeled as follows. The environment is passed onto *A.6.1.77* and then it first checks for अचि. Since only this is carried as anuvṛtti, once this condition evaluates to 'true', the rule notifies all rules registered with it. It waits for any of its descendants to claim eligibility. If no rule claims eligibility then the rule checks for additional condition of इकः and if the condition evaluates to 'true', it claims the eligibility. In case of conflict between *A.6.1.77* and *A.6.1.101*, the desired rule is *A.6.1.101*, but A.6.1.101 is a descendant of *A.6.1.77*, so it will claim its eligibility and hence *A.6.1.77* will be blocked from claiming its eligibility as per our system. From these examples it is evident that the system, without even using any extra conflict resolution techniques, can resolve the conflicts here. These examples demonstrate that the system internalizes rule *A.1.4.2* विप्रतिषेधे परं कार्यम्. Now consider the case when there is a conflict between the rules *A.7.3.111* and *A.7.1.73*. Here both the rules belong to domain of aṅga, i.e., rule *A.6.4.1*. Neither of the rules inherit from the other, and hence both of them are at sibling level and have a common parent at rule *A.6.4.1*. In such a case where both the rules claim eligibility, the conflict which is at the level of rule *A.6.4.1*, needs to be resolved. The desired rule in such a case is *A.7.1.73*. The important point to be noted here is that though the system internalizes the concept of *A.1.4.2*, it still does not claim *A.7.3.111* as the winner rule. Instead the system will pass both the rules to the conflict resolution method coded as a separate function in the system, once it reaches the object for the rule *A.6.4.1*. We will be trying out different conflict resolution methods mentioned in authentic texts and will be reporting the accuracy of each method or various combinations of these methods on different data samples. Other conflict cases discussed in Scharf (2010) are not discussed here as those fall into one of the scenarios already discussed.

# 7 Evaluation Results

## 7.1 Evaluation Framework

For the evaluation, we have implemented the entire apatya section in the Aṣṭādhyāyi. The apatya section consists of rules from *A.4.1.92 to A.4.176* which deal with the affixation rules for stems that need to be used along with semantic sense of apatyam (with its subtypes gotra, yuvām) or 'the descendant of' semantic relation. For the proper execution of these rules, we were required to implement other rules that come in the multilevel inheritance hierarchy like the pratyayādhikāra rule *A.4.1.83* and its exceptions as well as those outside the Taddhita section like vṛddhi, guṇa and other associated rules. We have selected 60 input cases that cover the whole span of 'apatya' section and obtained the output after affixation from the system.

A web based survey interface was used for the human judgment experiment.[1] We divided our inputs into sets of 3, each set having 20 input cases. We then reached out to the experts from Sanskrit linguistics to participate by evaluating one or more of these sets. A total of five experts from the linguistics and Sanskrit computational linguistics fields participated in the evaluation. For a given input case, the following details were provided to the experts: the input string, conditions and assumptions that were required for the correct derivation, winner rule number and the sūtra contents, candidate rules which had conflicts and finally the output form. The experts were asked to do a binary evaluation of the correctness of output, based on the input and other constraints provided. In case of difference of opinion among the experts, we took the majority opinion as the truth value after weighing in the remarks provided by all the experts.

Figure 8: Snapshot of the evaluation framework

The results of the evaluation are shown in Table 2. From among the 60 input cases, a total of 10 cases were evaluated as incorrect. As already discussed in Section 4.6, the system has internalized vipratiṣedha as its default conflict resolution mechanism. The outputs were obtained with no other conflict resolution mechanism being employed. 10 cases resulted in wrong output. We later implemented the specificity hierarchy for conflict resolution and then obtained the outputs again for the same set of input cases. This time, the number of incorrect output cases were reduced to 4. We will be discussing some of the wrong output cases in Section 7.2.

### 7.2 Analysis of Wrong Cases and Other Special Cases.

■ For inputs उत्स ङस् अपत्यम्, दिति ङस् अपत्यम्, the rules that got applied were *A.4.1.95 and A.4.1.122* respectively which essentially look for the phonological properties. The correct rules to be applied for the instances are *A.4.1.86 and A.4.1.85* respectively. Though these rules have the same topical heads, they are specified before the current winner rules in

---

[1] The evaluation URLs are :- Set 1 - http://goo.gl/forms/Lj5z3UzUr9, Set 2 - http://goo.gl/forms/2O31D6gF83, Set 3 - http://goo.gl/forms/goWgwMnYct

Table 2: Evaluation results for the experimental setup

(a) Details about the experimental setup

| Number of Input cases | 60 |
|---|---|
| Evaluators Participated | 5 |
| Inputs per evaluator | 20 |
| Numebr of Correct cases | 50 |
| Number of Wrong output (with no external conflict resolution) | 10 |
| Number of wrong outputs (with specificity hierarchy for conflict resolution) | 4 |

(b) Accuracy of the evaluation results

| Evaluation | Accuracy | Error |
|---|---|---|
| With no external conflict resolution | 83.33 | 16.67 |
| With specificity hierarchy | 93.33 | 6.67 |

Aṣṭādhyāyi, and hence by principle of vipratiṣedha, the former got preference over the desired ones. But if we observe they are more specific than the current winner rules. The correct rules have specificity at the morphological level because the rules mention some stems, while the current winner rules have specificity only at the phonological level. Please note that all the mentioned rules (winner and desired) carry entities that bear semantic specificity. But, since all the rules have that notion of semantic specificity, the effect from this notion is nullified here.

∎ For inputs गर्ग ङस् गोत्र, कपि ङस् गोत्र, the rules that got applied are *A.4.1.151 and A.4.1.122*, instead of the rules *A.4.1.105 and A.4.1.107* respectively. Any rule in apatya section, or for that matter any rule under the 'arthādhikāra' has a semantic specificity by default, which is the semantic sense. But in this case, the rules *A.4.1.105 and 107* bear a finer level of semantic specificity of gotra, while the current applied rules deal with the general case of apatya. Please note that gotra is a specific sub-type of apatya. So whenever the intended semantic sense is gotra, and if there exists a rule that is applicable and has gotra specification, then it should be applied if conflicting with a rule of apatya specification.

∎ For the input पितृष्वसृ ङस् अपत्यम्, ideally two output cases should appear, पितृष्वसृ ङस् छण् and पितृष्वसृ ङस् ढक्. It is not rare in Taddhita derivation to see the applicability of multiple affixes for a single input case. But here the second output is the result of the rule *A.4.1.133* ढकि लोपः. The rule states that the final ऋ in पितृष्वसृ will be elided. Now no other rule mentions about affixation of ढक् for the given input case. While implementing the rule *A.4.1.133*, an assumption is made that the at first affix ढक् should be introduced and then the elision operation should be performed. The basis of this assumption comes from the argument by many of the scholars that Pāṇini, by specifying about the condition in *A.4.1.133* implies that the affix ढक् should be introduced (Sharma, 2002). It is argued that for the sake of brevity, Pāṇini explicitly did not mention the rule, but if the affixation is not implied then there is no purpose for the rule.[2]

∎ In case of rule *A.4.1.122*, the rule is applicable, specifically for a nominal that has an 'i' as its last phonological entity but not by virtue of इञ् affix attached to them. Such rules cannot be triggered unless and until one has prior knowledge about the affixes that the word-form obtained during its formation. For this purpose while automating, we should either keep a list of words beforehand, or else there should be some mechanism to store the extra information. Our system currently stores a list of stems to tackle the issue as we will be taking text inputs from the user. But what is more significant is that our 'śabdarūpa' object provides the functionality to store all those extra information like the 'it' marker ञ् which otherwise will not be present in the final word-form. This implies that when we reuse a word form for derivation from our own system, we utilise those extra information which are stored in the data structure.

---

[2] Our implementation has followed this assumption.

# 8 Discussion and Conclusion

## 8.1 Bottlenecks in Automating the Astadhyayai

Our attempt was to automate the Taddhita section in Astadhyayai which primarily deals with affixation of derivational nouns. It is evident from the discussions so far, that Pāṇini uses a rich set of linguistic features in formulating rules and conditions to check for, before affixation. From the discussion about specificty hiererchy, we can see that the entire Aṣṭādhyāyi has used variety of linguistic features ranging from phonetic to semantic features. Along with such features, Pāṇini occasionally used the intention of the speaker as a condition to be satisfied as well. For example, if we consider the rule *4-1-147* गोत्रस्त्रियाः कुत्सने ण च , the term कुत्सन signifies 'reproach'. It is the intention of speaker whether or not to address someone by referring through the 'descendant of' relation in a disrespectful way. Such an aspect cannot be captured beforehand. So while implementing the automated system, we either had to generate all the alternatives, whenever the condition to be satisfied is an intention, or else ask for human intervention to clarify the intention. We proceeded with the later method as the former strictly does not adhere to the intended output derivation of the grammar.

In Section 7.2, the first case can raise a question about our assumption in the specificity hierarchy. The desired rule for the instance discussed is *A.4.1.86* and the conflicting rule is *A.4.1.95*. Now if we look into the rule contents, we can find that rule *A.4.1.86* does not come under any 'arthādhikāra', and only comes under a 'pratyayādhikāra' and *A.4.1.95* has an 'arthādhikāra'. It can be argued that *A.4.1.95* has a higher priority as per specificity hierarchy as it has a semantic condition, but the desired rule does not have a semantic condition, but only a morphological condition. These observations motivated us for the existing assumption. The linguistically motivated counter argument is that if we go by the former argument, then the rule *A.4.1.86* will never find its applicability. From the implementation aspect of view, the default affix rule *A.4.1.83* and its exceptions which include *A.4.1.86* as well, get to apply the prefix after a suitable semantic relation is found, which are later rules like *A.4.1.92*, which is in fact a semantic condition and this nullifies the effect. The assumption still needs to be validated in other similar rule conflict cases outside of Taddhita section.

In Section 6, we have talked in length about how the system has internalized the rule *A.1.4.2*. Among different type of rules that Pāṇini had formualted in his treatise, *A.1.4.2* is a metarule, that describes about how other rules are to be interpreted. There are many such metarules in Aṣṭādhyāyi like *A.1.1.66, A.1.1.49* etc. In our system we have internalised those concepts and directly applied those in our implementation logic. So in our implementation the metarules are assumed to be known and the metarules are followed, but there is no explicit entity defined for the same. Further research is required to code meta-rules as separate entities, to have their presence stated explicitly.

## 8.2 Conclusions and Future directions

The proposed schema, which is primarily aimed at the automated generation of derivative nouns, or what is called as Taddhita section, adopts a completely object oriented approach in modeling the system. To the best of the authors' knowledge, the system is first of its kind that focuses specifically on modeling Taddhita section of Aṣṭādhyāyī. The system not only focuses on generation of correct final forms but also in producing the correct sequence of rules applied for the entire derivation process. This approach doubles the system as a tool for pedagogy, where a learner can use the system to learn about the derivation process by trying out nominal of his choice, rather than restricting himself to a fixed set of textbook examples or a set of stored samples. As already discussed, the system facilitates in adopting and trying out various conflict resolution methods for rule selection, which can be programmed separately and then invoked from the system. As there is no consensus over one correct set of resolution methods, the proposed system can be used as a validation tool for verifying the accuracy of different conflict resolution methods that have been mentioned in various scholarly works. In fact, we have not yet mentioned about one of the major areas of application for the proposed system. Since the system is essentially generating nouns (or adjectives) from other nouns (or adjectives) based on semantic relations, the system can be used to obtain relations between different nominals in an automated way. This kind of information can be used to supplement available lexical resources like IndoWordNet for sanskrit (Bhattacharyya, 2010) and also to provide supplementary information in dictionary entries for lexical databases as described in Huet (2004).

Our system forms rule groups as Directed Acyclic Graphs (DAG), analogous to as Subbanna and Varakhedi (2009) suggests. But our notion of topical rule groups arises from the notion of topicality by virtue of anuvṛtti, while Subbanna and Varakhedi (2009) attempts to form the topical rule groups from general-exception rule concepts. The approaches in group formation need to be studied further, in order to comment on the commonality of the DAGs so formed, but Subbanna and Varakhedi (2009) do not talk about the attempts to model their concept in implementation. As each rule class implements precisely what is mentioned in the rule, and the implied portion is inherited by virtue of anuvṛtti from the parent class, the system can keep a trace of all the parent rules that have acted upon the environment, due to triggering of a particular rule. This gives the learner a much more detailed view of triggering of rules. The environment data structure is also an object based implementation, environment mostly stores properties and does not have methods or functions that model any aspect of Aṣṭādhyāyī. The functions or methods implemented in environment are mostly for programmatic conveniences or adding functionality to the system front-end. This makes the set up of environment comparable to the set up as in Scharf et al. (2015) which is primarily an XML representation. The conversion of Objects to XML schema is a well addressed task in IT World (Jain and Thakur, 2002). Hence one can also think of developing a schema to transfer the data entities used in one system to the other, as Scharf et al. (2015) tries to model the entire Aṣṭādhyāyī.

## Acknowledgments

We would like to extend our gratitude towards **Dr. Gérard Huet, Dr. Peter Scharf and Dr. Amba Kulkarni** for their support and valuable inputs from the discussions we had related to the work. We would also like to thank the experts, **Ms. Sukhada, Ms. Anuja Ajotikar, Ms. Tanuja Ajotikar, Mr. Pavankumar Satuluri and Mr. P Sanjeev** for participating in the system evaluation we had conducted and also for sharing their valuable comments.

## References

Saroja Bhate. 1989. *Panini's Taddhita rules*. University of Poona, Pune.

Pushpak Bhattacharyya. 2010. Indowordnet. In *proceedings of LREC-10*. Citeseer.

George Cardona. 1965. On translating and formalizing pāṇinian rules. In *Journal of the Oriental Institute of Baroda*, volume 14, pages 306–14.

George Cardona. 1969. Studies in indian grammarians i: the method of description reflected in the śivasūtras. In *Transactions of the American Philosophical Society*, pages 3–48. JSTOR.

George Cardona. 1997. Panini: His work and its traditions vol 1. In *Background and Introduction. 2nd ed.* Motilal Banarsidass.

George Cardona. 2009. On the structure of pāṇini's system. In *Sanskrit Computational Linguistics, First and Second International Symposia, Rocquencourt, France*, pages 1–32. Springer.

Ashwini Deo. 2007. Derivational morphology in inheritance-based lexica: Insights from pāṇini. In *Lingua*, volume 117.1, pages 175–201. Elsevier.

Erich Gamma, Richard Helm, Ralph Johnson, and John Vlissides. 1994. *Design Patterns: Elements of Reusable Object-Oriented Software*. Addison Wesley.

Pawan Goyal and Gérard Huet. 2013. Completeness analysis of a sanskrit reader. In *Proceedings, 5th International Symposium on Sanskrit Computational Linguistics. DK Printworld (P) Ltd*, pages 130–171.

Pawan Goyal, Amba Kulkarni, and Laxmidhar Behera. 2009. Computer simulation of aṣṭādhyāyī: Some insights. In *Sanskrit Computational Linguistics, First and Second International Symposia, Rocquencourt, France*, pages 139–161. Springer.

Gérard Huet. 2004. Design of a lexical database for sanskrit. In *Proceedings of the Workshop on Enhancing and Using Electronic Dictionaries*, pages 8–14. Association for Computational Linguistics.

Malcolm D. Hyman. 2009. From pāṇinian sandhi to finite state calculus. In *Sanskrit Computational Linguistics, First and Second International Symposia, Rocquencourt, France*, pages 253–265. Springer.

S. Jain and S. Thakur. 2002. Xml to object translation, June 13. US Patent App. 09/755,501.

Girish Nath Jha and Sudhir K. Mishra. 2009. Semantic processing in pāṇini's kāraka system. In *Sanskrit Computational Linguistics, First and Second International Symposia, Rocquencourt, France*, pages 239–252. Springer.

Girish Nath Jha, Muktanand Agrawal, Subash, Sudhir K. Mishra, Diwakar Mani, Diwakar Mishra, Manji Bhadra, and Surjit K. Singh. 2009. Inflectional morphology analyzer for sanskrit. In *Sanskrit Computational Linguistics, First and Second International Symposia, Rocquencourt, France*, pages 219–238. Springer.

S. D. johshi and J. A. F. Roodbergen. 1987. On siddha, asiddha and sthĀnivat. In *Annals of the Bhandarkar Oriental Research Institute*, volume 68, pages 541–549. Bhandarkar Oriental Research Institute.

Malhar Kulkarni. 2009. Phonological overgeneration in pāṇinian system. In *Sanskrit Computational Linguistics, First and Second International Symposia, Rocquencourt, France*, pages 306–319. Springer.

Anand Mishra. 2009. Simulating the pāṇinian system of sanskrit grammar. In *Sanskrit Computational Linguistics, First and Second International Symposia, Rocquencourt, France*, pages 127–138. Springer.

Anand Mishra. 2010. Modelling aṣṭādhyāyī: An approach based on the methodology of ancillary disciplines (vedanga). In *Sanskrit Computational Linguistics, Fourth International Symposium, Delhi, India*, pages 239–258. Springer.

Gerald Penn and Paul Kiparsky. 2012. On pāṇini and the generative capacity of contextualized replacement systems. In *Proceedings of COLING 2012: Posters*, pages 943–950.

Pawankumar Satuluri and Amba Kulkarni. 2014. Extra linguistic information needed for automatic generation of sanskrit compounds: A study. In *The recent developments in Sanskrit Computational Linguistics', at SALA-30, Hyderabad*.

Peter Scharf, Pawan Goyal, Anuja Ajotika, and Tanuja Ajotikar. 2015. Voice, preverb, and transitivity restrictions in sanskrit verb use. In *Sanskrit Syntax, Selected papers presented at the seminar on sanskrit syntax and discouse structures*, pages 157–202.

Peter Scharf. 2009. Modeling pāṇinian grammar. In *Sanskrit Computational Linguistics, First and Second International Symposia, Rocquencourt, France*, pages 95–126. Springer.

Peter M Scharf. 2010. Rule selection in the aṣṭādhyāyī, or is pāṇini's grammar mechanistic? In *Studies in Sanskrit Grammars: Proceedings of the Vyakarana Section of the 14th World Sanskrit Conference*.

Rama Nath Sharma. 2002. *The Aṣṭādhyāyi of Pāṇini - Vol.1 : Introduction to the Aṣṭādhyāyi as a Grammatical Device*. Munshiram Manoharlal Publishers Pvt. Ltd., New Delhi.

J. Frits Staal. 1965. Context-sensitive rules in pāṇini. In *Foundations of Language 1*, pages 63–72.

Sridhar Subbanna and Shrinivasa Varakhedi. 2009. Computational structure of the aṣṭādhyāyī and conflict resolution techniques. In *Sanskrit Computational Linguistics, Third International Symposium, Hyderabad, India*, pages 56–65. Springer.

Sridhar Subbanna and Shrinivasa Varakhedi. 2010. Asiddhatva principle in computational model of aṣṭādhyāyī. In *Sanskrit Computational Linguistics, Fourth International Symposium, Delhi, India*, pages 231–238. Springer.

Constantin Szallies. 1997. On using the observer design pattern. *XP-002323533,(Aug. 21, 1997)*, 9.